\documentclass[conference]{IEEEtran}
\IEEEoverridecommandlockouts
\usepackage{cite}
\usepackage{amsmath,amssymb,amsfonts}
\usepackage{amsthm}
\usepackage{mathtools} 
\usepackage[ruled,vlined,linesnumbered]{algorithm2e} 
\usepackage{algorithmic} 
\usepackage{graphicx}
\usepackage{textcomp}
\usepackage[table]{xcolor}
\usepackage{nicematrix}
\usepackage{tabularray}
\usepackage{xcolor}
\usepackage{booktabs,siunitx}
\usepackage{soul} 
\usepackage{url}
\usepackage{tikz}
\usetikzlibrary{arrows}
\usepackage{stackengine}
\usetikzlibrary{fit}
\usetikzlibrary{backgrounds}
\usetikzlibrary{shapes.arrows}
\def\BibTeX{{\rm B\kern-.05em{\sc i\kern-.025em b}\kern-.08em
    T\kern-.1667em\lower.7ex\hbox{E}\kern-.125emX}}

\newcommand{\vxacut}[1]{ }
\newcommand{\vxapara}[1]{\noindent{\bf{#1}}}

\newcommand{\hlred}[1]{{\sethlcolor{red}\hl{#1}}}
\newcommand{\needswork}[1]{\hl{needs work:}[#1]}

\newcommand{\masanote}[0]{\hlred{Masa Note:}}

\newcommand{\masacut}[1]{ }
\newtheorem{clm}{Claim}

\newcommand{\bsx}[0]{{\bf x}}

\newcommand{\bsy}[0]{{\bf y}}

\newcommand{\bstheta}[0]{{\boldsymbol{\theta}}}
\newcommand{\glatent}[0]{{\bf h}}
\newcommand{\mpnfunc}[0]{g}

\DeclareMathOperator*{\argmin}{argmin}

\DeclareMathOperator*{\cost}{cost}
\newcommand{\gset}[0]{\mathcal{G}}
\newcommand{\gvset}[0]{\mathcal{V}}
\newcommand{\geset}[0]{\mathcal{E}}
\newcommand{\gyset}[0]{\mathcal{Y}}
\newcommand{\gpath}[0]{p} 

\newcommand{\gpathset}[0]{\mathcal{P}}
\newcommand{\gpathsetgt}[0]{\mathcal{P}^{+}}
\newcommand{\gpathsetopt}[0]{\mathcal{P}^{*}}

\newcommand{\gwcost}[0]{c}
\newcommand{\gwset}[0]{\{\gwcost_e\}_{e=1}^{|\geset|}}  
\newcommand{\superplus}[0]{\scriptscriptstyle{+}}  
\newcommand{\superminus}[0]{\scriptscriptstyle{-}}
\newcommand{\superplusminus}[0]{\scriptscriptstyle{+/-}}

\newcommand{\gnodefeats}[0]{\gyset}
\newcommand{\targetobs}[0]{\bsy}
\newcommand{\targetstate}[0]{\bsx}
\newcommand{\targettrack}[0]{\gpath}
\newcommand{\alltracks}[0]{\gpathset}
\newcommand{\gnnparams}[0]{\bstheta}
\newcommand{\targetobsspatial}[0]{\targetobs^{\mathbf{(s)}}}
\newcommand{\targetobsreid}[0]{\targetobs^{\mathbf{(id)}}}

\vxacut{
https://fusion2024.org/
https://fusion2024.org/call-for-papers/
}

\begin{document}

\title{SSP-GNN: Learning to Track \\
via Bilevel Optimization\\
\thanks{$^{*}$Corresponding author}
}

\author{\IEEEauthorblockN{Griffin Golias}
\IEEEauthorblockA{\textit{Applied Physics Laboratory} \\
\textit{University of Washington}\\
Seattle, WA 98105 USA \\
goliagri@uw.edu}
\and
\IEEEauthorblockN{Masa Nakura-Fan}
\IEEEauthorblockA{\textit{Paul G. Allen School of Computer Science} \\
\textit{University of Washington}\\
Seattle, WA 98105 USA \\
mnakura@uw.edu}
\and
\IEEEauthorblockN{Vitaly Ablavsky$^{*}$}
\IEEEauthorblockA{\textit{Applied Physics Laboratory} \\
\textit{University of Washington}\\
Seattle, WA 98105 USA \\
vxa@uw.edu } 
}

\maketitle

\begin{abstract}
We propose a graph-based tracking formulation for multi-object tracking (MOT) where target detections contain kinematic information and re-identification features (attributes). Our method applies a successive shortest paths (SSP) algorithm to a tracking graph defined over a batch of frames. The edge costs in this tracking graph are computed via a message-passing network, a graph neural network (GNN) variant. The parameters of the GNN, and hence, the tracker, are learned end-to-end on a training set of example ground-truth tracks and detections. Specifically, learning takes the form of bilevel optimization guided by our novel loss function. We evaluate our algorithm on simulated scenarios to understand its sensitivity to scenario aspects and model hyperparameters. Across varied scenario complexities, our method compares favorably to a strong baseline.

\vxacut{
Multi-object tracking remains a challenging problem even when target detections contain features (attributes) that could be exploited for data association. One challenge is optimally exploiting both kinematic information and high-dimensional features and doing so in a computationally-efficient manner. 
}

\vxacut{
Multi-object tracking remains a challenging problem even when targets can be distinguished by their features (attributes). In particular, one needs to solve data association in a way that is both optimal and computationally efficient. We propose a trainable formulation for batch-based data association taking into account kinematics and target features. Our formulation that takes into account targets' kinematics and feature information via graph neural network (GNN) to compute edges scores of a tracking graph. These edge scores guide a successive-shortest-path (SSP) algorithm to optimally find trajectories in the tracking graph. We conduct experiments to understand the advantages of our formulation and the effect of hyper-parameters on the tracking solution.
}

\end{abstract}

\section{Introduction}
Multi-target and multi-object tracking (MOT) is often a necessary step in larger systems that address real-world challenges. Examples include tracking pedestrians in the context of autonomous driving, tracking animals and birds to understand environmental factors, and tracking players in team sports to analyze plays.

MOT remains a challenging problem even in scenarios where detections contain features (attributes) that could potentially help associate across target identities and distinguish targets from clutter. One of the challenges in feature-aided tracking is to exploit feature vectors of arbitrary (possibly high) dimensions. Another challenge is to reason jointly and optimally about attributes and target dynamics.

\vxacut{
Machine-learning methods, such as deep neural nets and graph neural nets (GNNs) have been shown to learn useful representations for decision tasks given high-dimensional inputs, but it's not clear how to best apply those ML methods to MOT.
}

Learning-based approaches to tracking~\cite{Xiang_learning_to_track_ICCV2015, Schulter_CVPR_2017} learn key components of a tracker (e.g., data association) directly from annotated training data.  Recently, approaches to neural message passing~\cite{pmlr-v70-gilmer17a} have been adopted by~\cite{Braso_2020_CVPR} to formulate tracking on a {\em detection graph} (where a detection graph is instantiated for a temporal window given all the measurements/detections).  However, predicting which pairs of detections should be linked without inferring complete tracks, while computationally efficient, may be suboptimal.

We propose an end-to-end bilevel formulation for learning MOT, where the inner optimization (tracking) is accomplished via a successive shortest paths (SSP) algorithm  on a {\em tracking graph}~\cite{k_best_paths_castanon1990, Zhang_CVPR_2008} with edge costs. Given these costs, SSP finds a solution that is guaranteed to be optimal. In our outer optimization problem we learn parameters of a function that computes the cost of each edge in the tracking graph given detections that include attributes. The edge-cost prediction function takes the form of a graph neural network (GNN).

Learning GNN parameters requires solving a bilevel optimization problem that is different from those proposed earlier, e.g.~\cite{Schulter_CVPR_2017, li_learning_tracking_2022}. Specifically, we derive a novel loss function defined with respect to SSP-computed tracks and the ground-truth tracks and an algorithm to learn the GNN parameters by gradient descent. At each iteration, GNN parameters are updated to increase the cost of incorrect tracks, effectively learning from (tracking) mistakes. 

The contributions of our paper are as follows:
\begin{itemize}
    \item A novel end-to-end learnable approach to graph-based tracking with lower computational cost than~\cite{Schulter_CVPR_2017, li_learning_tracking_2022}.
    \item A novel use of SSP for inner optimization, guaranteeing an optimal solution satisfying tracking constraints. \vxacut{, and is known~\cite{berclaz_pami2011} to be more efficient than its linear-programming (LP) alternatives}
    \item Quantitative analysis of our algorithm on a diverse set of synthetic scenarios and comparison with a strong baseline that relies on GNN but does not employ global path optimization.
\end{itemize}

\vxacut{
The parameters of the GNN are learned iteratively by correcting mistakes made by the tracker.

Our formulation is attractive from the computational-complexity standpoint Successive-shortest paths (SSP) algorithms have been shown to be more efficient~\cite{berclaz_pami2011} than their LP counterparts.

Our contribution is to derive an end-to-end formulation that draws on ideas from~\cite{pmlr-v70-gilmer17a} but also includes an optimization problem on a tracking graphs. We claim that this yields the complete tracking solution without the need for post-processing heuristics.

We validate our approach...We show that our learned function generalizes to all temporal windows and all tracking graphs where targets’ features come from a
distribution “similar” to the distribution of features seen during
training.
}

\section{Related Work}\label{sec:related_work}
\vxapara{Learning approaches to tracking} In many domains, using machine-learning tools to find optimal parameters for a tracking algorithm from data have been shown to outperform hand-crafting and hand-tuning an algorithm. In the computer vision domain, learning-based approaches to multi-object tracking have taken many different forms. \cite{Xiang_learning_to_track_ICCV2015} demonstrated a Markov decision process (MDP) capable of initiating and terminating target tracks; parameters of this MDP were learned from data. In \cite{rnn_tracking_milan2017} recurrent neural networks were used in an end-to-end framework. Specifically, in~\cite{rnn_tracking_milan2017} a target's state was maintained by a dedicated recurrent neural network (RNN); assignment of measurements to target tracks and track creation was estimated via a long short-term memory (LSTM) neural network. A {\em transformer} neural network was used in~\cite{Meinhardt_2022_CVPR} as a unified formulation for detecting (new) targets, extending tracks, and creating new tracks.

\vxapara{Graph-based tracking via classical methods} Application of graph-based and linear-programming (LP) methods to multi-target tracking has a long history, e.g.,~\cite{k_best_paths_castanon1990}, and includes recent attempts to combine graph-based tracking with multi-hypothesis tracking (MHT)~\cite{coraluppi_carthel_2019multiple}.
Graph-based tracking has been shown to be effective in computer vision where it is often employed with a tracking-by-detection paradigm where tracking is reduced to the grouping of detections (bounding boxes) proposed by a (target class specific) object detector. 

An early graph-based method of~\cite{Zhang_CVPR_2008} proposed the general form of a {\em tracking graph} and solved for an unknown number of targets via repeated invocation of min-cost flow. The computational complexity was reduced in~\cite{Pirsiavash_CVPR_2011} by introducing {\em augmenting paths}, thus extending an existing tracking solution to include an additional target instead of starting from scratch. While~\cite{Zhang_CVPR_2008} and~\cite{Pirsiavash_CVPR_2011} performed tracking in the image plane, tracking on the ground plane from multiple camera views via the K-shortest-paths (KSP) algorithm was proposed in~\cite{berclaz_pami2011}. Approaches mentioned thus far assumed that target dynamics are Markovian and targets move independently of each other. An approach to extend LP-based tracking to take into account long-term (group) behavior was proposed in~\cite{non_markovian_Maksai_ICCV_2017}. During training, their method alternated between learning behavior patterns and fitting trajectories to data given the observations and priors given by the learned patterns.

\vxapara{Graph-based tracking via neural methods} A significant limitation of early graph-based approaches~\cite{Zhang_CVPR_2008,Pirsiavash_CVPR_2011, berclaz_pami2011} is the requirement to hand-craft scalar costs on tracking graph edges. However, in many real-world scenarios where we wish to apply feature-aided tracking, measurements (detections) in addition to containing positional information also include (high-dimensional) vectors of attributes, such as re-identification (ReID) features. Neural networks could learn a mapping from these high-dimensional attributes to scalar edge weights, but deriving an end-to-end learnable method is nontrivial. Among the first attempts to apply neural methods to graph-based tracking, the approach of~\cite{Schulter_CVPR_2017} stands out for formulating learning and tracking as bilevel~\cite{intro_to_bilevel_opt_zhang2023} optimization. Specifically, in~\cite{Schulter_CVPR_2017} the high-level optimization problem solved for the neural network parameters, while the lower-level problem was a constrained LP that solved a tracking-by-detection problem over a temporal window of pre-defined length. The bilevel approach of~\cite{Schulter_CVPR_2017} was revisited in~\cite{li_learning_tracking_2022}, but instead of relying on the implicit function theorem to deal with the inner LP problem, the latter used KKT conditions to compute the gradient at the lower problem's optimum.

Recent advances in neural methods for graph-structured problems, such as graph neural networks (GNNs)~\cite{defferard2016} led to advances in graph-based tracking. Inspired by the neural message-passing on graphs method of~\cite{pmlr-v70-gilmer17a}, the authors of~\cite{Braso_2020_CVPR} proposed to learn a function to estimate edge probabilities in a tracking graph. This approach was extended in~\cite{Cetintas_braso_2023_CVPR} to jointly solve for short-term and long-term tracking. While the approaches of~\cite{Braso_2020_CVPR, Cetintas_braso_2023_CVPR} require a batch of frames, and thus operate with a constant lag, approaches of~\cite{Wang2020_GNNDetTrk, Meinhardt_2022_CVPR} require only the current frame to manage tracks. The ideas of~\cite{pmlr-v70-gilmer17a} were adopted in~\cite{liangnebp_fusion2022} to develop a neural-enhanced message-passing data association algorithm. 

\vxapara{Neural approaches to optimization} Learning-based approaches to MOT mentioned thus far solve the underlying assignment problem using classical methods e.g., LP. An approach to solve quadratic optimization as a layer in a neural network was proposed in~\cite{optnet_pmlr-v70-amos17a}. In~\cite{sinkhorn_papakis2020gcnnmatch}, the matrix of association costs was updated using Sinkhorn-Knopp's algorithm~\cite{sinkhorn1967concerning} to satisfy MOT constraints. A differentiable graph matching layer was proposed in~\cite{learnable_graph_matching_He_CVPR2021} and used to associate graphs formed on detections across time frames.

\vxapara{Our contribution} We draw on~\cite{defferard2016, pmlr-v70-gilmer17a} and~\cite{Braso_2020_CVPR} to formulate a GNN that infers edge costs on a graph structure, but unlike~\cite{Braso_2020_CVPR} our inner optimization with SSP ensures that track predictions are globally optimal and satisfy constraints. We demonstrate in Section~\ref{sec:experiments} that our method compares favorably to a GNN-based method without such inner optimization.
Our method is similar to that of~\cite{li_learning_tracking_2022} in forming a bilevel problem during training, but we show in Section~\ref{sec:comp_complexity} that
our inner optimization with SSP has lower computational complexity.

\vxacut{
We draw on the ideas from~\cite{defferard2016, pmlr-v70-gilmer17a} and~\cite{Braso_2020_CVPR} to formulate a GNN that infers edge costs on a graph structure. However, unlike~\cite{Braso_2020_CVPR} our predictions are guaranteed to form valid tracks because we rely on SSP for inference. Our method is similar to that of~\cite{li_learning_tracking_2022} in forming a bilevel problem during training, but our inner optimization problem is different to the benefit of computational efficiency.
}

\vxacut{
Intro to bilevel optimization~\cite{colson_overview_bilevel_2007} ~\cite{intro_to_bilevel_opt_zhang2023} bilevel nonsmooth~\cite{bilevel_opt_nonsmooth_ochs2015}

Not relevant?
LeNoach's feature-aided MHT~\cite{lenoach_feature-aided_2021}

Bras\'{o}~\cite{Braso_2020_CVPR}, latest Bras\'{o}~\cite{Cetintas_braso_2023_CVPR} , Li's linear programming with MLP~\cite{li_learning_tracking_2022}, Gilmer's neural message passing~\cite{pmlr-v70-gilmer17a}, , Liang's neural enhanced belief prop (FUSION 2022)~\cite{liangnebp_fusion2022},  Schulter's deep network flow for MOT~\cite{Schulter_CVPR_2017}. GNN tracking online~\cite{Wang2020_GNNDetTrk}

}
\section{Approach}
Let $\targetstate$ denote target state and let $\targetobs$ denote measurement (detection). A detection $\targetobs_{j,t}$ at time $t$ may originate from one of the targets (if present) or from clutter (i.e., it's a false alarm). 

We adopt a sliding-window approach, where tracking is performed on a batch of data frames over a temporal window of length $T$. In a complete system, separate logic is used to produce an output that takes into account the tracking solution from the previous temporal window. Figure~\ref{fig:sys_diagram} shows the system diagram of our approach.

\tikzstyle{int}=[fill, fill=cyan!20, minimum size=1em]
\tikzstyle{int1}=[fill, fill=gray!10, minimum size=0em]
\tikzstyle{int2}=[fill, fill=orange!10, fill opacity=0, text opacity=1, minimum size=0em]
\tikzstyle{init} = [pin edge={to-,thin,black}]
\tikzstyle{input} = [coordinate]
\tikzstyle{output} = [coordinate]
\tikzstyle{background1}=[rectangle, fill=cyan!20, inner sep=0.3cm, opacity=1, rounded corners=5mm]
\tikzstyle{background2}=[rectangle, fill=orange!20, inner sep=0.3cm, opacity=0.5, rounded corners=5mm]
\tikzset{MyArrow/.style={single arrow, draw=black, fill=gray!20, minimum width=10mm, minimum height=61mm, inner sep=0mm, single arrow head extend=1mm}}
\tikzset{BackArrow/.style={single arrow, draw=black, fill=gray!20, minimum width=10mm, minimum height=50mm, inner sep=0mm, single arrow head extend=1mm}}

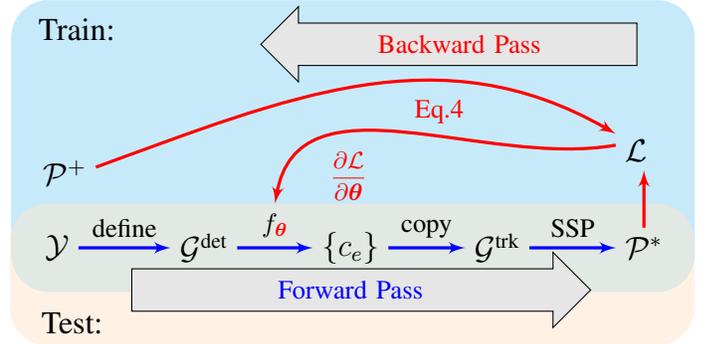
\begin{figure}
\centering
\begin{tikzpicture}[node distance=1.95cm,auto,>=latex']
    \node [int2] (y_tr) [] {\large $\gyset$};
    \node [coordinate] (p_coord) [above of=y_tr, node distance=1cm] {};
    \node [int] (p_gt) [right of=p_coord, node distance=0.1cm] {\large$\gpathsetgt$};
    \node [int2] (g_obs_tr) [right of=y_tr] {\large$\gset^{\textrm{det}}$};
    \node [int2] (costs_tr) [right of=g_obs_tr] {\large$\{\gwcost_e\}$};
    \node [int2] (g_trk_tr) [right of=costs_tr] {\large$\gset^{\textrm{trk}}$};
    \node [int2] (p_opt_tr) [right of=g_trk_tr] {\large\large$\gpathsetopt$};
    \node [int] (loss) [above of=p_opt_tr, node distance=1.3cm] {\large$\mathcal{L}\phantom{*}$};
    \node [coordinate] (tr_coord) [above of=y_tr, node distance=2.9cm] {};
    \node [int] (train_label) [right of=tr_coord, node distance=0.26cm] {\large Train:};

    \path[blue, ->, very thick] (y_tr) edge node [black] {define} (g_obs_tr);
    \path[blue, ->, very thick] (g_obs_tr) edge node [name=eq1, black] {$f_{\textcolor{red}{\gnnparams}}$} (costs_tr);
    \path[blue, ->, very thick] (costs_tr) edge node [black] {copy} (g_trk_tr);
    \path[blue, ->, very thick] (g_trk_tr) edge node [black] {SSP} (p_opt_tr);
    \path[red, ->, very thick] (p_opt_tr) edge node {} (loss);
    \draw[red, ->, very thick] (p_gt.10) to [out=20, in=150] node [name=p_gt_to_L] {} (loss);
    \draw[red, ->, very thick] (loss.170) to [out=190, in=80] node[near end] {\small $\dfrac{\partial \mathcal{L}}{\partial \gnnparams}$} node [name=eq5, swap] {Eq.\ref{eq:ssp_gnn_chain_rule}} (eq1.north);

    \node [coordinate] (forward_arrow_start) [below of=y_tr, node distance=1cm] {};
    \node [coordinate] (forward_arrow_end) [below of=loss, node distance=2cm] {};
    \node [coordinate] (backward_arrow_start) [above of=loss, node distance=1.7cm] {};
    \node [coordinate] (backward_arrow_end) [above of=eq1, node distance=2.7cm] {};
    \path (forward_arrow_start) -- (forward_arrow_end) node[midway, MyArrow, name=forward_arr, text=blue] {Forward Pass};
    \path (backward_arrow_start) -- (backward_arrow_end) node[midway, BackArrow, name=backward_arr, shape border rotate=180, text=red] {Backward Pass};

    \node [coordinate] (tst_coord) [below of=tr_coord, node distance=3.9cm] {};
    \node [coordinate] (box_coord) [left of=tst_coord, node distance=0.33cm] {};
    \node [int2] (test_label) [right of=tst_coord, node distance=0.2cm] {\large Test:};

    \begin{pgfonlayer}{background}
        \node [background1, fit=(p_gt)(eq5)(loss)(backward_arr)(tr_coord)(y_tr)(p_opt_tr)] {};
        \node [background2, fit=(y_tr)(p_opt_tr)(tst_coord)(box_coord)] {};
    \end{pgfonlayer}
\end{tikzpicture}
\caption{System diagram of our learnable method. $f_\gnnparams$ corresponds to Eq.\ref{eq:edge_cost}, explained in detail by Algorithm \ref{alg:message_passing}. In our bi-level formulation (Eq.\ref{eq:bilevel_opt}), learning $\gnnparams$ is the outer objective, while SSP is the graph-based inner optimization that explicitly produces globally optimal tracks $\gpathsetopt$ in $\gset^{\textrm{trk}}$. GNN parameters $\gnnparams$ are updated in the backward pass through Eq.\ref{eq:ssp_gnn_chain_rule}.}
\label{fig:sys_diagram}
\end{figure}

\vxacut{\node [int] (g_trk_copied) [below of=costs_tst, node distance = 0.75cm] {$\gset^{\textrm{trk}}$};
    \node [int] (g_obs_tst2) [left of=g_trk_copied, node distance = 2cm] {$\gset^{\textrm{det}}$};
    \node [int] (path_opt_tst) [right of=g_trk_copied, node distance = 2cm] {$\gpathsetopt$};
    \path[blue, ->] (g_obs_tst2) edge node [swap]{transfer} (g_trk_copied);
    \path[blue, ->] (g_trk_copied) edge node [swap, name=ssp] {SSP} (path_opt_tst);}

\subsection{Graph-based formulation}  

Within the batch of $T$ data frames, we solve the tracking problem by casting it in a graph-based formulation~\cite{Zhang_CVPR_2008, Pirsiavash_CVPR_2011}. Given a temporal window $t \in \{1,\ldots,T\}$ our objective is to group $\{\targetobs_{j,t}\}$ into a set $\alltracks \doteq \{\targettrack_k\}$ of paths in a {\em tracking graph} $\gset^{\textrm{trk}}$ with positive and negative edge costs. The set $\alltracks$ may be empty if the tracker doesn't predict any tracks.


\subsection{The detection graph and the tracking graph}

\begin{figure}[h]
\begin{center}
\includegraphics[width=0.99\linewidth, trim={0.99in 8.00in 0.17in 0.05in},clip]{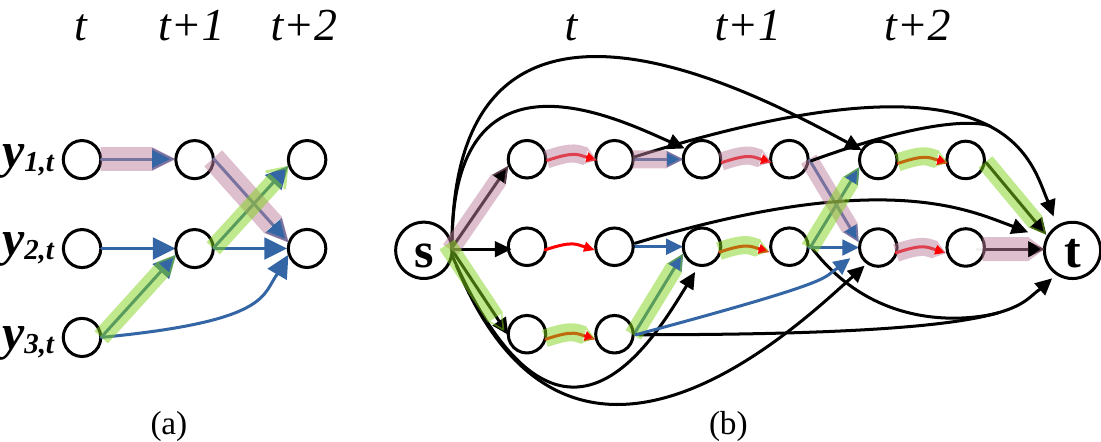}
    \caption{Our formulation employs two types of graphs: (a) {\em detection graph} $\gset^{\textrm{det}}$ constructed from measurements in a given temporal window; (b) its corresponding {\em tracking graph} $\gset^{\textrm{trk}}$.}
    \label{fig:xog_tg}
\end{center}
\end{figure}

Given detections $\gyset \doteq \{\targetobs_{j,t} \}_{t=1}^{T}$ in a temporal window with $T$ frames, we construct a detection graph $\gset^{\textrm{det}}$. Nodes in $\gset^{\textrm{det}}$  are in one-to-one correspondence with detections in $\gyset$. An edge $(j,j')$ exists between $\targetobs_{j,t}$ and $\targetobs_{j',t'}$ if $t < t' \leq t + \Delta$ and the location components of $\targetobs_{j,t}$ and $\targetobs_{j',t'}$ satisfy {\em gating} constraints. The use of $\Delta > 1$ is standard~\cite{Schulter_CVPR_2017, li_learning_tracking_2022} to handle short-duration obscuration (occlusion) of a target. In Fig.~\ref{fig:xog_tg}(a) we show an example of $\gset^{\textrm{det}}$ for $T=3$ and $\Delta=2$. Two paths that satisfy node-disjoint constraints are shown.

Given $\gset^{\textrm{det}}$, we define its companion {\em tracking graph} $\gset^{\textrm{trk}}$, whose topology follows~\cite{k_best_paths_castanon1990, Zhang_CVPR_2008}. In particular, every node in $\gset^{\textrm{det}}$ is represented by a pair of ``twin'' nodes in $\gset^{\textrm{trk}}$. Furthermore, $\gset^{\textrm{trk}}$ contains source ${\bf s}$ and terminal ${\bf t}$ nodes with edges from ${\bf s}$ and to ${\bf t}$ defined in a way as to allow every detection to either correspond to a track's birth or its termination. Fig.~\ref{fig:xog_tg}(b) shows the tracking graph $\gset^{\textrm{trk}}$ derived from its $\gset^{\textrm{det}}$; the paths (tracks) in $\gset^{\textrm{trk}}$ originate in ${\bf s}$, pass through ``twin'' nodes and terminate in ${\bf t}$.

\vxacut{
In Figure~\ref{fig:xog_tg} we show two types of graphs used by our algorithm $\gset^{\textrm{det}}$ and $\gset^{\textrm{trk}}$ 
}

\vxacut{
Solving the tracking problem on $G$ can be accomplished in several ways. [discuss LP, min-cost flow, KSP/SSP]
Resulting in a set of tracks $\alltracks \doteq \{\targettrack \}$
}

\vxacut{\subsection{Computing edge-costs of a tracking graph via MPN}}

\vxapara{Computing edge-costs of a tracking graph via MPN} Let $f_\gnnparams$ be a function, parameterized by $\gnnparams$ for computing edge costs in a graph $\gset^{\mathrm{det}} = (\gvset, \geset, \gyset)$ 
\begin{equation} \label{eq:edge_cost}
    f_\gnnparams : \gset^{\mathrm{det}} \rightarrow \gwset\,.
\end{equation}
Our $f_\gnnparams$ takes the form of a message-passing network (MPN)~\cite{pmlr-v70-gilmer17a}, a specialization of a graph neural net (GNN)~\cite{defferard2016}. We adopt the ideas of~\cite{pmlr-v70-gilmer17a} and ~\cite{pmlr-v80-kipf18a} as implemented in~\cite{Braso_2020_CVPR}.

\vxacut{
Let $h_v^t$ be the attributes of vertex $v$ at timestep $t$ and $e_{v,w}^t$ be the attributes of the edge between $v$ and $w$ at time t. Let $E$ denote the set of edges and $\alpha$ be the ReLU activation function.
\begin{equation}
h_v^{t+1} = \alpha(W_1(\alpha(W_0 h^t_v),\sum_{e_{v,w} \in E}e_{v,w}^t))
\end{equation}
\begin{equation}
e_{v,w}^{t+1} = \alpha(W_4(\alpha(W_2 e^t_{v,w}),\alpha(W_3(h_v^t, h_w^t))))
\end{equation}
Where $W_0, \dots W_4$ are learned weight matrices.

The basic update step for undirected graph G where $v$ is a vertex, $e_{vw}$ is an edge in $G$ is given by:
\begin{equation}
    m_v^{t+1} = \sum_{w \in N(v)} M_t(h_v^t,h_w^t,e_{vw})
\end{equation}
\begin{equation}
    h_v^{t+1} = U_t(h_v^t,m_v^{t+1})
\end{equation}
where $M_t$ is a learned function, $U_t$ is some update function, $N(v)$ is the neighborhood of v, and $h_v^{t}$ is the hidden state of vertex $v$ at time $t$.
}

Let $L$ be an integer representing the number of message passing layers (iterations), $\geset$ be the edge set, and $\gvset$ be the vertex set of $\gset^{\mathrm{det}}$. Let $\glatent^{(l)}_{(i,j)}$ be the embedding of edge $(i,j) \in \geset$ at iteration $l$ and similarly $\glatent^{(l)}_z$ be the embedding of the vertex $z \in \gvset$ at iteration $l$. We assume without loss of generality that node $j$ is ahead of node $i$ in time. 
\begin{algorithm}[hbt!]
\caption{Computing edge costs in $\gset^{\textrm{det}}$}
\label{alg:message_passing}
Initialize $\glatent^{(0)}_{(i,j)}$ and  $\glatent^{(0)}_{z}$ for all $(i,j) \in \geset, z \in \gvset$ \\
$l \gets 1$ \\
\While{$l \le L$}{
    \For{$(i,j) \in \geset$}{
        $\glatent^{(l)}_{(i,j)} = \mpnfunc_e\left([\glatent_i^{(l-1)}, \glatent_j^{(l-1)},\glatent_{(i,j)}^{(l-1)}]\right)$ \\
        $m_{j,(i,j)}^{(l)} = \mpnfunc_v^{\mathrm{past}}\left([\glatent_j^{(l-1)},\glatent^{(l)}_{(i,j)}]\right)$ \\
        
        $m_{i,(i,j)}^{(l)} = \mpnfunc_v^{\mathrm{fut}}\left([\glatent_i^{(l-1)},\glatent^{(l)}_{(i,j)}]\right)$ \\
    }
    \For{$z \in \gvset$}{
        $\glatent_{z, \mathrm{past}}^{(l)} =\sum_{(i,z) \in \geset}{m^{(l)}_{z,(i,z)}}$ \\
        $\glatent_{z, \mathrm{fut}}^{(l)} =\sum_{(z,j) \in \geset}{m^{(l)}_{z,(z,j)}}$ \\    
        $\glatent_z^{(l)} = \mpnfunc_v  \left([\glatent_{z,\mathrm{past}}^{(l)}, \glatent_{z,\mathrm{fut}}^{(l)}] \right)$ \\ 
    } 
    $l \gets l+1$ \\ 
} 
\For{$(i,j) \in \geset$}{
    $\gwcost_{(i,j)} = g_{e}^{\mathrm{readout}}(\glatent^{(L)}_{(i,j)}, \glatent^{(L)}_{(j,i)})$ \\
}

\textbf{Output:} $\gwset$ 
\end{algorithm}

Messages are computed using the functions $g_e(\cdot, \cdot, \cdot)$ and $g_v(\cdot, \cdot)$ which take the form of learnable multi-layer perceptrons. In our implementation, as suggested by~\cite{Braso_2020_CVPR}, $g_v(\cdot, \cdot)$ consists of two independently parameterized functions $g_v^{\mathrm{past}}(\cdot, \cdot), g_v^{\mathrm{fut}}(\cdot, \cdot)$ which are specialized for forward and backward in time messages respectively. The readout function $g_e^{\mathrm{readout}}(\cdot, \cdot)$, also a learnable MLP, then computes the edge cost by aggregating the final attributes of the forward and backward in time edges between each pair of nodes. The parameters of these four functions comprise the vector $\gnnparams$.

\subsection{End-to-end learning for tracking}

\vxacut{
We seek $h_\theta$ that generalizes to all temporal windows and all tracking graphs where targets' features come from a distribution ``similar'' to the distribution of features seen during training. 
}

\vxapara{Optimization criterion}  
To derive our algorithm to learn $f_\gnnparams$ we need a differentiable optimization criterion (a loss function). The loss function should be defined with respect to the ground-truth tracks (paths) $\gpathsetgt$ and the predicted ones $\gpathsetopt$.

Given $\gset^{\mathrm{trk}} = (\gvset, \geset, \gyset)$ with edge costs $\gwset$ we rewrite the edge costs as a real-valued function $c(\cdot)$. For a track (path) $p \in \gset^{\mathrm{trk}}$, we define (with slight abuse of notation) $c(p) = \sum_{e \in p}{c(e)} + c^{\mathrm{en}} + c^{\mathrm{ex}}$, where $c^{\mathrm{en}} > 0, c^{\mathrm{ex}} > 0$ are entrance and exit costs. For a set of paths $\gpathset$, we define $c(\gpathset) = \sum_{p \in \gpathset}{c(p)}$.

We define the {\em set} loss $\mathcal{L}$ for two sets of paths $\gpathsetgt, \gpathsetopt$ as 
\begin{equation}
\mathcal{L}(\gpathsetgt, \gpathsetopt) = \ell_1(\gpathsetgt, \gpathsetopt) + \ell_2(\gpathsetgt)\,.     
\end{equation}
The first term is defined $\ell_1(\gpathsetgt, \gpathsetopt) = \max(0, c(\gpathsetgt) - c(\gpathsetopt))$ and the second $\ell_2(\gpathsetgt) = \sum_{p\in \gpathsetgt}{\max(0,c(p))} $.

The following claim follows immediately from the properties of an SSP algorithm.
\begin{clm} Let $\gpathsetopt$ be the estimated set of paths. Then $c(\gpathsetopt) \le 0$ and for all $p \in \gpathsetopt$, $c(p) \le 0$.
\end{clm}

\noindent The next claim confirms that optimizing $\mathcal{L}$ converges to $\gpathsetgt$.
\begin{clm}
    Suppose that the cost of each unique set of mutually disjoint paths in the tracking graph is unique. Then $\ell_1(\gpathsetgt,\gpathsetopt) = 0$ if and only if $\gpathsetgt = \gpathsetopt$ as sets. If these equivalent conditions hold, $\mathcal{L}(\gpathsetgt,\gpathsetopt) = 0$ as well.
\end{clm}
The first part follows from the fact that $\gpathsetopt$ is the global optimum, but so is $\gpathsetgt$, by definition. The second part follows by showing that $\ell_2(\gpathsetgt) = 0$, and hence $\mathcal{L}$ is zero.

\vxapara{Bilevel optimization to learn MPN parameters} We formulate the following bilevel optimization problem~\cite{colson_overview_bilevel_2007, intro_to_bilevel_opt_zhang2023, bilevel_opt_nonsmooth_ochs2015} to find the optimal set of GNN parameters given ground-truth paths $\gpathsetgt$ :
\begin{equation}\label{eq:bilevel_opt}
\begin{split}
    \gnnparams^{*} = &\argmin_\gnnparams \mathcal{L} (\gpathsetgt, \gpathsetopt) \\
    &\textrm{s.t.}, \alltracks^{*} = \mathsf{track\_by\_ssp}(\gset,\gwset)
\end{split}    
\end{equation}
%
The problem defined in Equation~\ref{eq:bilevel_opt} is non-trivial: the outer objective is non-linear, while the inner objective is stated in terms of a  constrained search for minimum-cost paths in a graph. 
\vxacut{Had the inner problem been stated as a relaxed LP as in~\cite{Schulter_CVPR_2017,li_learning_tracking_2022} applying the chain rule would have yielded
\begin{equation}\label{eq:LP_chain_rule}
\dfrac{\partial \mathcal{L}}{\partial \gnnparams }=\dfrac{\partial \mathcal{L}}{\partial \bsx^{*}} \cdot \underbrace{\dfrac{\partial \bsx^{*}}{\partial {\bf c}}}_{\substack{\text{optimum} \\ \text{w.r.t costs}}} \cdot \underbrace{\dfrac{\partial {\bf c}}{\partial \gnnparams }}_{\substack{\text{costs} \\ \text{w.r.t MPN}}}
\end{equation}
where $\bsx$ defines which edges in the tracking graph are ``active'', so that $\alltracks^{*}$ would be encoded in the $\bsx^{*}$; the vector of per-edge costs ${\bf c}$ would be the output of an MPN.  
Since we are using a path-following algorithm for inner optimization of Equation~\ref{eq:bilevel_opt}, our loss function is different.
} 
Applying the chain rule to our loss, yields 
\begin{equation}\label{eq:ssp_gnn_chain_rule}
\dfrac{\partial \mathcal{L}}
{\partial \gnnparams }=\dfrac{\partial \mathcal{L}}{\partial {\bf \gwcost}} \Big |_{\alltracks^{*}} \dfrac{\partial {\bf \gwcost}}{\partial \gnnparams} \;,
\end{equation}
where ${\bf \gwcost} \doteq \gwset$. 
\vxacut{
Comparing Equation~\ref{eq:LP_chain_rule} with Equation~\ref{eq:ssp_gnn_chain_rule}, the biggest difference is the absence of the $\partial \bsx^{*}/\partial {\bf c}$ term, which ``guides'' the inner optimization toward the ground-truth solution. 
}

\vxacut{
We will demonstrate in the experiments section that learning $f_\gnnparams$ using Equation~\ref{eq:bilevel_opt} is a stable training algorithm.
Here we present a simple extension that leads to fast convergence in practice from a randomly-initialized $\gnnparams^{\mathrm{init}}$. We initially  train the MPN   using a surrogate loss and without bilevel formulation to obtain $\hat{\gnnparams}$  
, then search for the optimal $\gnnparams^{*}$ following gradient descent via Equation~\ref{eq:bilevel_opt}.
}

In practice, we implement optimization of $\gnnparams$ in Equation~\ref{eq:bilevel_opt} via a two-stage learning algorithm. In Stage I, inspired by~\cite{oh_sastry_mcmcda_2009} which introduced MCMC data association {\em moves}, we ``perturb'' $\gpathsetgt$ to pre-compute a fixed-size set of negative examples $\gpathset^{-}$. Generating $\gpathset^{-}$ does not require running the SSP algorithm.

Let $ \gpath^{\superminus} \in \gpathset^{-}$ be one of the paths (tracks) obtained via perturbation, and let $ \gpath^{\superplus} \in  \gpathsetgt $ be a ground-truth path. In general, given $\gnnparams^{*}$, it is not the case that every $\gpath^{\superminus}$ has a higher cost than every  $\gpath^{\superplus}$. Nevertheless for Stage I we define
\begin{equation}
\ell^{\scriptscriptstyle{+-}}(\gpath^{\superplus},\gpath^{\superminus}) = \mathsf{max}(0,\cost(\gpath^{\superplus}) - \cost(\gpath^{\superminus} )+ \gamma)
\end{equation}
for some margin $\gamma \geq 0$. And we can extend this to the entire set $\gpathset^{-}$ of all the paths obtained by perturbing $\gpathsetgt$. Optimizing $\ell^{\scriptscriptstyle{+-}}$ leads to $\hat{\gnnparams}$ which serves as a ``warm start'' for Stage II which implements Equation~\ref{eq:bilevel_opt}.

\vxacut{
Mention that our optimization is general: (a) can be any path-following on graphs (b) applies to extended-object tracking or point targets. 
}

Our end-to-end learning approach is defined in Algorithm~\ref{alg:ssp_gnn}; extending it to multiple training graphs is straightforward. We will demonstrate in the experiments section that learning $f_\gnnparams$ this way yields a stable training algorithm. 

  \begin{algorithm}
  	\caption{Training SSP-GNN } 
    \label{alg:ssp_gnn}
	\textbf{Inputs} :  \\ 
		$\gset = (\gvset,\geset,\gnodefeats)$ \quad (training) graph with node features $\gnodefeats $ \\
		$\mathcal{P}^{\superplus} \triangleq \mathcal{P}^{\mathsf{GT}} $ \quad ground-truth paths (tracks) in $\gset$ \\
  Stage I parameters $M, N_1^{\mathsf{max}}$ ; Stage II parameter $\epsilon \geq 0$\\
		$f_{\bstheta} : \gset  \rightarrow \gwset , \ \gwcost_e \in \mathbb{R}  $
	
	\textbf{// Stage I (bootstrapping)}  \\
		// $\mathcal{P}^{\superminus} \triangleq \{ \gpath_1^{\superminus}, \mathellipsis, \gpath_M^{\superminus} \} $ \\
  	$\gpathset^{\superminus} \leftarrow \mathsf{perturb\_paths}(\gset,\gpathsetgt, M )  $ \\
   $\bstheta \leftarrow \mathsf{init\_gnn\_weights}() $ \\
	\While{$ \mathsf{itr} < N_1^{\mathsf{max}} $ }
       {   \vspace{0.2em} 
		$\gwset \leftarrow f_\bstheta(\gset) $ \\
		// compute loss $\ell^{\scriptscriptstyle{+-}}(\gpathsetgt, \gpathset^{\superminus})$; update $\bstheta$ \\
	}

  		
	\textbf{// Stage II} 	\\
		\While { {\tt True} }
		{  \vspace{0.2em} 
		$\gwset \leftarrow f_\bstheta(\gset) $ \\
		$\gpathsetopt \leftarrow \mathsf{track\_by\_ssp}(\gset,\gwset) $ \\
         \If {$\mathcal{L(\gpathsetgt, \gpathsetopt)} < \epsilon$} {{\tt break}} 
        update $\bstheta$ 
	    } 
    \textbf{Output:} $f_{\gnnparams}$ 
  \end{algorithm}

\subsection{Tracking with a trained model}  
Once training is complete, the model weights are frozen and inference is performed without a backward pass, i.e. the bottom arrow of Fig. \ref{fig:sys_diagram}. Our experiments evaluate model performance over individual tracking windows. Our method can be extended to long tracking sequences by splitting them into smaller temporal windows and applying standard methods for re-initializing and stitching tracks across overlapping windows \cite{Braso_2020_CVPR}, \cite{li_learning_tracking_2022}.

\subsection{Computational Complexity}\label{sec:comp_complexity}
\vxacut{
[vxa $\rightarrow$ all: This is a sketch; please revise/rewrite!] In our analysis, to help with a fair comparison We focus only on the inner optimization.
As shown in~\cite{TODO}, computational complexity of a K-shortest-paths algorithm grows with size of $\gyset$ as $O(k \ldots \log \ldots)$ where $n$ is the size of $\mathcal{G}^{\mathrm{trk}}$ and $k$ is the number of paths/tracks estimated by the algorithm. Note that $n$ is a function of the number of nodes {\em and} the number of edges, and therefore depends on the size of the temporal window {\em} and the number of targets and false alarms (i.e., the complexity of the scenario).

By contrast, as shown in~\cite{TODO} computational complexity of LP grows as $O(n^2)$ and QP...is given by... [Cite and discuss OptNet paper?]}

In our analysis, to provide a fair comparison between our proposed algorithm and related methods we focus only on forward inference of the inner optimization problem used to compute $\mathcal{P}^\star$ from $\mathcal{G}^\mathrm{trk}$. In our implementation, this is solved using SSP, specifically the $k$-shortest-paths algorithm. As shown in \cite{berclaz_pami2011}, this algorithm has a worst-case time complexity of $\mathcal{O}(k(m + n\cdot \mathrm{log}\:n))$, where $k$ is the number of paths/tracks found by the algorithm and $m$ and $n$ refer to the number of edges and nodes, respectively, contained in the graph $\mathcal{G}^\mathrm{trk}$. The size and topology of $\mathcal{G}^\mathrm{trk}$ depends on $\mathcal{Y}$, and therefore is a function of both the size of the temporal window and the number of targets and false alarms.

In contrast, linear programming (LP)-based approaches to the min-cost flow problem have an average-case polynomial time complexity in the size of $\mathcal{G}^\mathrm{trk}$ \cite{Ahuja1993NetworkFT}. This also applies to more recent approaches presented in \cite{Schulter_CVPR_2017} and \cite{li_learning_tracking_2022}, which embed the LP solution as a differentiable layer within a multi-layer perceptron.\footnote{Note that \cite{li_learning_tracking_2022} transforms the LP problem to a regularized quadratic program; however, the transformed problem has the same time complexity.} The proposed SSP-GNN algorithm blends the edge-cost expressivity of learning-based methods with the reduced time complexity of SSP methods.

\section{Experiments}\label{sec:experiments}
We conduct extensive experiments to demonstrate the advantages of our method and to assess the impact of a variety of model and scenario attributes on performance.

\subsection{Train and Test Scenarios}
Our synthetic data is generated using Stone Soup~\footnote{\url{https://github.com/dstl/Stone-Soup/}}, an open-source framework. We generate diverse scenarios with multiple targets using constant velocity and Ornstein-Uhlenbeck motion models. When generating detections, we fix the detection probability $(P_D)$ at 1.0 except in Table~\ref{tab:p_d_exps} which investigates the effects of an imperfect detector, and we fix measurement error standard deviation at $0.2$. We refer to target features as re-identification (ReID) as is common in computer vision. However, our approach can generalize to other features and sensing modalities, e.g.,~\cite{lenoach_feature-aided_2021}. Our ReID features are drawn from a mixture of Gaussians (MoG) distributions. We control the ``strength'' of ReID features by changing the amount of overlap between the foreground and background distributions and define categories in terms of KL divergence: {\em very weak} (0.02 nats/0.029 bits), {\em weak} (0.5 nats/0.721 bits), {\em moderate} (3.125 nats/4.508 bits), and {\em strong} (12.5 nats/18.034 bits).

\subsection{Implementation Details} 
 Given a pair of detections $\targetobs_{i,t}, \targetobs_{i',t'}$ with $t \neq t'$ we follow~\cite{li_learning_tracking_2022} to encode their spatio-temporal constraint into our latent representation. We denote the kinematic (spatial) component of a detection as $\targetobsspatial_{i,t}$, $\targetobsspatial_{i',t'}$, the ReID component as $\targetobsreid_{i,t}$, $\targetobsreid_{i',t'}$. For a node $v$ in $\gset$, we define $\glatent^{(0)}_{v} = [\targetobsspatial_v, \targetobsreid_v]$. For an edge $e$ between the two detections in $\gset$, we define $\glatent^{(0)}_{e} = [t'- t, \targetobsspatial_{i'} - \targetobsspatial_{i}, \|\targetobsreid_{i'} - \targetobsreid_{i}\|]$. We use 2D scenarios and 2D ReID features, i.e., $\targetobsspatial \in \mathbb{R}^2$ and $\targetobsreid \in \mathbb{R}^2$.

\vxapara{Tracking graph and inference} We instantiate $f_\gnnparams$ and learn parameters $\gnnparams$ on the detection graph $\gset^{\textrm{det}}$ as shown in Fig.~\ref{fig:xog_tg}(a). Then, $f_\gnnparams(\gset^{\textrm{det}})$ assigns scalar costs to the edges in $\gset^{\textrm{det}}$. We copy these edge costs from $\gset^{\textrm{det}}$ to $\gset^{\textrm{trk}}$ by associating each edge $e_{i,j} \in \gset^{\textrm{det}}$ with the edge between the pair of twin nodes associated with $i$ and the pair of twin nodes associated with $j$ in $\gset^{\textrm{trk}}$. We assign a cost of 0 to the edges within the twin nodes, shown as red arrows in Figure~\ref{fig:xog_tg}(b). This formulation allows us to use the GNN-computed $\gwset$ as edge costs for the SSP algorithm to optimize over $\gset^{\textrm{trk}}$. Detection and tracking graphs are represented via NetworkX~\footnote{\url{https://networkx.org/}} except for GNN learning and inference, where the detection graph is converted to a Deep Graph Library (DGL)~\cite{wang2019dgl} object. 

\vxacut{For the inner optimization problem, we adapt an SSP solver, similar to~\cite{berclaz_pami2011}.
Detection and tracking graphs are represented via NetworkX~\footnote{\url{https://networkx.org/}} but for instantiating GNN and parameter learning the detection graph is converted to a Deep Graph Library (DGL)~\cite{wang2019dgl} object. }

\begin{figure}[h]
\begin{center}
\includegraphics[width=0.89\linewidth,trim={0in 0in 0.35in 0.55in},clip]{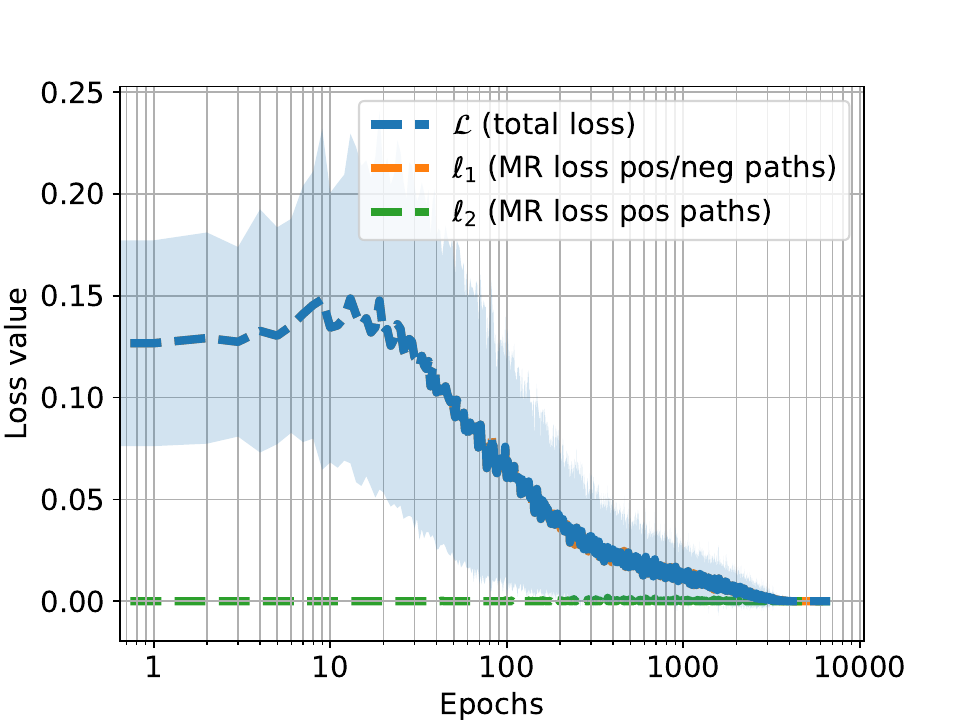}
    \caption{The average Stage II loss values across 20 SSP-GNN models with randomly initialized weights. As the number of epochs increases, the average loss converges to zero with decreasing standard deviation (shaded). }
    \label{fig:train_loss}
\end{center}
\end{figure}

\vxapara{Training} We use the Adam optimizer~\cite{kingma2017adam} with a learning rate of 0.001. Fig.~\ref{fig:train_loss} demonstrates a typical loss curve across 20 re-initialized training trials on the same train set. In practice we observe that $\ell_1$ tends to dominate total loss $\mathcal{L}$ because (a) our Stage I bootstrapping guarantees that $\ell_2 \approx 0$ at epoch 0 and (b) $\ell_2$ captures the more trivial problem of simply keeping the ground truth paths sufficiently negative in cost, while $\ell_1$ addresses the more difficult problem of predicting the ground truth paths over competing sets of paths.

\subsection{Evaluation Metrics} 
Our method is inspired by~\cite{Braso_2020_CVPR, li_learning_tracking_2022}, etc., which were developed for tracking objects in video sequences. These methods tend to be evaluated with respect to Multiple Object Tracking Accuracy (MOTA)~\cite{bernardin_clear_mot}. While our method is applicable to tracking extended targets in video sequences, we generate point observations in our Stone Soup simulation. We therefore report the MOTA metrics calculated with respect to the point observations for all of our experiments. We also report GOSPA \cite{Rahmathullah_2017} and SIAP\cite{siap_votruba2001} metrics when comparing our model's performance to a filter-based tracker provided by Stone Soup.

Following~\cite{bernardin_clear_mot}, MOTA is defined as: 
\begin{equation}
\text{MOTA} = 1 - \frac{\sum_t |\text{FP}_t| + |\text{FN}_t| + |\text{IDS}_t|}{\sum_t |\text{gtDets}_t|}\,,
\end{equation}
where $t$ is the time step, $|\text{FP}_t|$ is the number of false positives, $|\text{FN}_t|$ is the number of false negatives, and $|\text{IDS}_t|$ is the number of identity switches. We report MOTA, as well as each constituent metric as ratios with the number of ground truth detections (rates). Since we control data generation, we can utilize ID-based matching between tracks and true detections instead of the traditional Hungarian matching algorithm typically used for computer vision applications. The ability to perform more straightforward ID-based matching for evaluation is one advantage of using synthetic data.

Both GOSPA and SIAP metrics are popular evaluation methods for comparing predicted tracks to continuous ground-truth object paths. GOSPA quantifies localization and cardinality errors, and SIAP completeness (C), ambiguity (A), spuriousness (S), and positional error, provide a more complete insight into the accuracy of associations between predicted tracks and ground truth objects. We report metrics summed over all time frames for GOSPA and averaged for SIAP.  

\subsection{Qualitative Demonstration}
\begin{figure}[h]
\centering
\includegraphics[scale=.385,trim={0.8in 0in 1.15in 0.5in},clip]{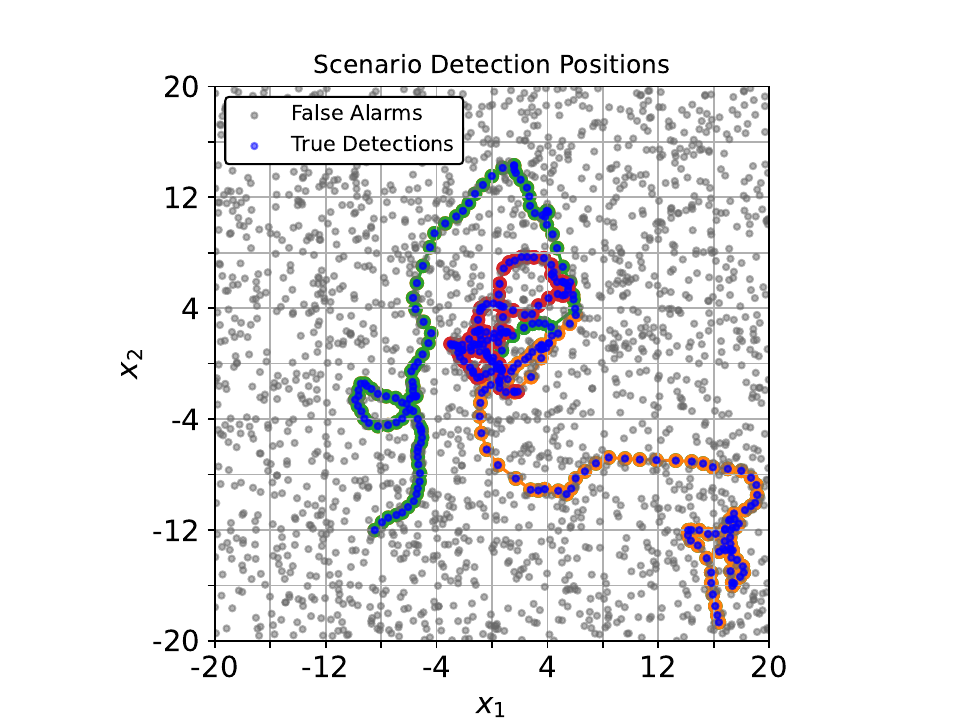}
\includegraphics[scale=.385,trim={0.8in 0in 1.15in 0.5in},clip]{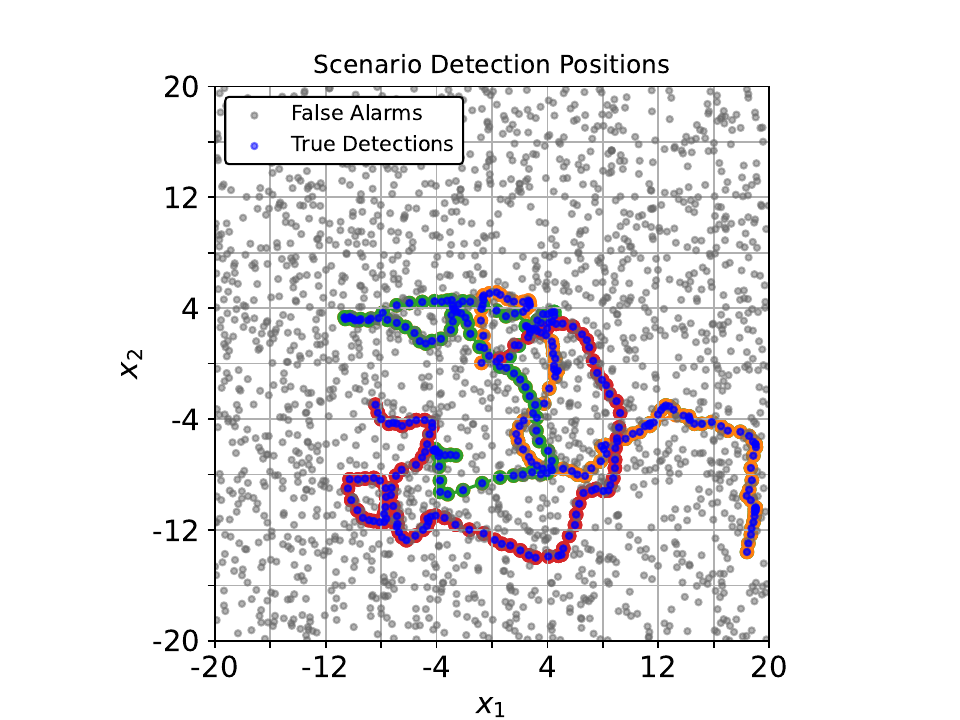}
\caption{Predicted tracks for the train (left) and test (right) scenario.} 
\label{fig:ssp_gnn_qualitative_1}
\end{figure}

For demonstration, in Figure~\ref{fig:ssp_gnn_qualitative_1} we show a sample SSP-GNN result trained on one scenario. Both the training and test scenarios last 100 timesteps and include three targets with randomly-generated lifespans following Ornstein-Uhlenbeck dynamics. Each scenario uses moderate strength ReID features and 20 uniformly distributed false alarms per frame. We observe that our algorithm generalizes well to the test scenario (the circled predicted tracks follow ground truth paths). The MOTA score is 0.993 for train scenario and 0.987 for the test scenario.

\vxacut{
\begin{table}[!htp]
\centering
\caption{MOTA metrics of Predicted Tracks for the sample train and test scenarios}
\sisetup{detect-all}
\NewDocumentCommand{\B}{}{\fontseries{b}\selectfont}
\begin{tabular}{
  @{}
  l
  S[table-format=1.2]
  S[table-format=1.2]
  S[table-format=1.2]
  S[table-format=1.2]
  S[table-format=1.2]
  @{}
}
\toprule
ReID Str. & {MOTA $\uparrow$} & {FP Rate $\downarrow$} & {FN Rate $\downarrow$} & {IDS Rate $\downarrow$} \\
\midrule
Train & 0.993 & 0 & 0 & 0.007 \\
Test & 0.987 & 0.003 & 0.003 & 0.007  \\
\bottomrule
\label{tab:ssp_gnn_qualitative_1_metrics}
\end{tabular}

\end{table}
}

\vxacut{
\hl{TODO} Generate figures like in LeNoach et al. FUSION 2021
instead of baseline vs. proposed, our figure will (a) results on training graph (b) results on test graph. Let's set $T=50$, number of GNN layers whatever makes for a good result. One target in clutter; target follows Ornstein-Uhlenbeck process or another linear transformation model, clutter is uniform and looks ``impressive''. Use strong ReID features. Note that visualizations we use internally for debugging purposes may not be the viz style for a conference publications. Try to replicate the look-and-feel for LeNoach's plots.
}

\subsection{Quantitative Evaluations} 
Since we are working with synthetic data, we can generate datasets of varying sizes and complexity. This enables isolating specific aspects of train and test scenarios to examine the effect on our model. However, since not all published methods report such analysis, and there is no standardized evaluation protocol, our results cannot be immediately compared to others reported in literature.

\begin{figure}[h]
\centering
\includegraphics[width=0.90\linewidth, trim={0in 0in 0in 0.175in},clip]{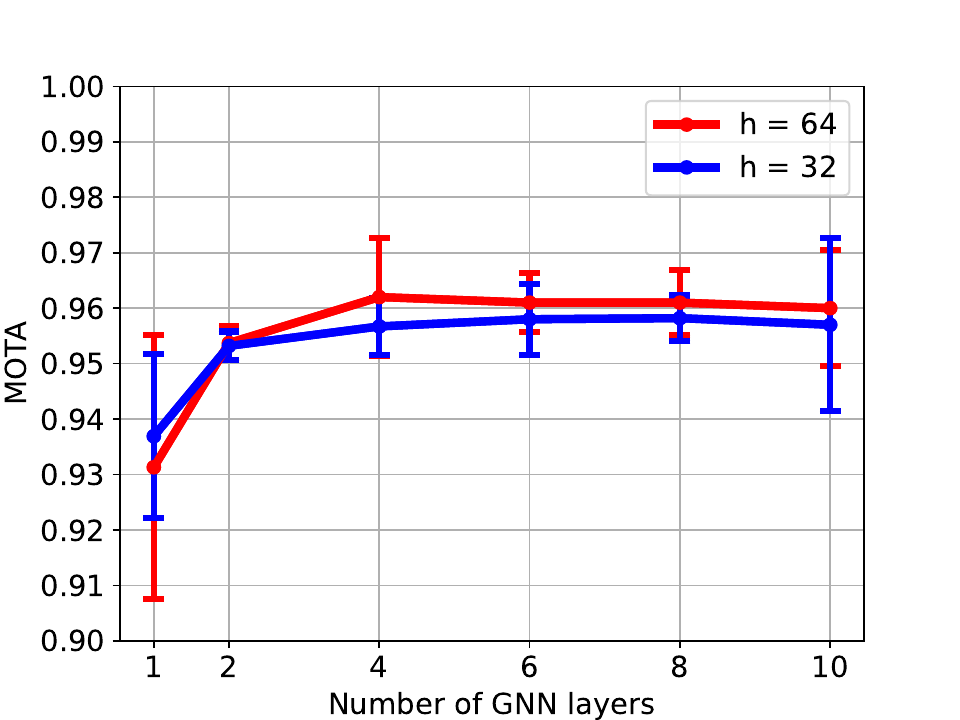}
\caption{Effect of the number of GNN message passing layers and hidden dimension size. Note that as the model capacity increases so does the tracking accuracy, until saturation.} \label{fig:eval_gnn_layers}
\end{figure}

\vxapara{Effect of GNN hyperparameters}
We first determine the optimal value of two hyperparameters used in configuring the graph neural network: number of message passing layers and size of the hidden layers $\glatent$ in all used MLPs. Figure~\ref{fig:eval_gnn_layers} shows the MOTA performance of the model on fixed train and test sets with variable number of GNN layers and hidden dimension size. The scenarios involve multiple ground truth tracks with constant velocity motion models, moderate strength ReID features, and 8 false alarms per frame in expectation.
\par
As the number of message passing layers in the graph neural network increases model performance also tends to increase until saturation is reached, at which point it largely levels out. Using a hidden dimension size of 64 produced the best results overall, and it achieves saturation at 4 GNN layers. Therefore, we use these hyperparameter values in all experiments going forward. When working with substantially different environments from our synthetic data, the optimal values for these hyperparameters would need to be adapted.

\vspace{-0in}

\vxapara{Comparison to baselines}
We compare our SSP-GNN tracker to an Extended Kalman Filter (EKF) based multi-object tracker and Edge-Belief GNN (based on~\cite{Braso_2020_CVPR}). The EKF tracker implementation is provided by the Stone Soup library. Although EKF only processes the kinematic component of measurements and is an online method (while ours is offline/batch-based), it establishes a reference point in our comparison. 

The Edge-Belief GNN algorithm~\cite{Braso_2020_CVPR} uses the same MPN architecture as our SSP-GNN, but instead of predicting edge costs, Edge-Belief GNN individually estimates the probability of an edge in the detection graph being part of a track or not. This is accomplished by modifying $g_e^{\mathrm{readout}}(\cdot, \cdot)$ to output sigmoid rather than logits. During inference, edge beliefs are rounded up or down to yield binary predictions, and a follow-on edge traversal heuristic ensures that the computed paths are node-disjoint. We observed that the Edge-Belief GNN tends to produce short (spurious) tracks; we added a post processing step which removes tracks shorter than a set length in order to strengthen the baseline (empirically we found minimum path threshold $=3$ was optimal). 

\vxacut{Scenarios are generated with 5 ground truth targets following constant velocity motion models with moderate ReID feature strength and an average of 8 false alarms per frame. These scenarios were chosen in part to enable the EKF tracker to perform competitively.}

\par
\begin{figure}[h]
    \centering
    \input{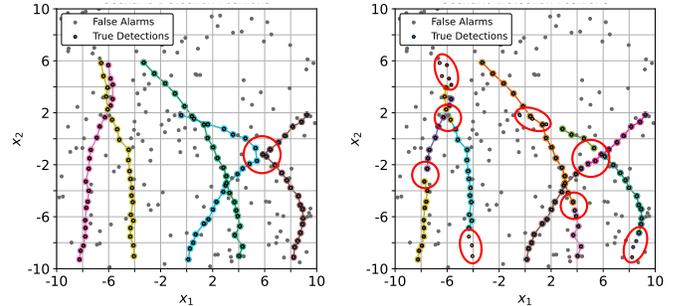}
\caption{SSP-GNN (left) and Edge-Belief GNN (right) algorithms trained and tested on identical sets. Small, colored dots with black outline are detection of true paths, while the larger circles linked together are algorithm generated tracks. Edge-Belief GNN tends to fragment tracks more often. Errors in the tracks are marked with red circles. Note false alarm density is $60\%$ higher than in Figure~\ref{fig:ssp_gnn_qualitative_1}.}
\label{fig:baseline_qualitative}   
\end{figure}

Figure~\ref{fig:baseline_qualitative} shows tracks produced by an SSP-GNN model and Edge-Belief GNN model on a test scenario. In this scenario, there are several instances where two tracks intersect. Our model correctly maintains target tracks on the left, while it swaps paths causing two identity switches on the right. Edge-belief GNN tends to fragment or deviate from tracks to nearby false alarms much more frequently than the SSP-GNN.  

 Table~\ref{tab:baseline-comparisons} compares Edge-Belief GNN, SSP-GNN, and EKF on similar scenarios. As inferred from the visual demonstration, the false negative rate is nearly twice that of SSP-GNN. Edge-Belief also has a substantially higher identity switch rate, indicating track fragmenting, but it does achieve a comparable false positive rate to our model.

Table~\ref{tab:edgebelief_vs_ssp} compares Edge-Belief GNN and SSP-GNN on similarly structured scenarios, but with increased difficulty due to false alarm rate and ReID feature weakness (the EKF tracker is omitted as ignoring ReID features makes it uncompetitive). SSP-GNN consistently outperforms Edge-Belief by a sizable margin, which is even more pronounced in harder regimes. 

\par

Because the Edge-Belief GNN baseline uses the same underlying message passing algorithm as SSP-GNN, we ascribe the observed performance increase to bilevel optimization. The SSP solver allows our model to properly weigh various trade-offs in the context of the full scenario. For instance it can weigh the cost of including a series of positive cost edges that would link otherwise disparate tracks versus the extra entrance/exit cost of inferring more tracks.

\vxacut{
\needswork{Remove enitrely???} Both online methods perform much better than EKF, which is to be expected given the advantages of offline inference and use of ReID features. [This sentence should probably go somewhere else earlier : GG]
}

\begin{table*}[!htp]
\centering
\caption{Comparison with baseline trackers. $\uparrow$ indicates higher is better. Parentheses next to metrics indicate a target value. }
\sisetup{detect-all}
\NewDocumentCommand{\B}{}{\fontseries{b}\selectfont}
\centering
\begin{tabular}{
  @{}
  l
  S[table-format=1.2]
  S[table-format=1.2]
  S[table-format=1.2]
  S[table-format=1.2]
  S[table-format=1.2]
  S[table-format=1.2]
  S[table-format=1.2]
  S[table-format=1.2]
  S[table-format=1.2]
  S[table-format=1.2]
  S[table-format=1.2]
  S[table-format=1.2]
  S[table-format=1.2]
  @{}
}
\toprule
Tracker & \multicolumn{11}{c}{Metrics} \\
\cmidrule(l){2-13}
& \multicolumn{4}{c}{MOTA} & \multicolumn{4}{c}{SIAP} & \multicolumn{4}{c}{GOSPA} \\
\cmidrule(lr){2-5} \cmidrule(lr){6-9} \cmidrule(lr){10-13}
& {MOTA $\uparrow$} & {FPR $\downarrow$} & {FNR $\downarrow$} & {IDS Rate $\downarrow$} & {C (1)} & {A (1)}& {S (0)} & {Pos. Err. (0)} & {Dist. $\downarrow$} & {Local. $\downarrow$} & {Missed $\downarrow$} & {False $\downarrow$}\\
\midrule
EKF &  &  & & & 0.945 & 1.065 & 0.055 & 0.164 & 23.130 & 14.130 & 2.750 & 6.250\\
Edge-Belief & \text{0.915$\pm$0.008} & \bf 0.018 & 0.041 & 0.027 & 0.961 & 1.009 & \bf0.008 & 0.129 & 8.772 & \bf 2.073 & 1.940 & \bf0.710\\
SSP-GNN (ours) & \text{\bf 0.942$\pm$0.050} & 0.022 & \bf 0.022 & \bf0.013 & \bf 0.977 & \bf1.000 & 0.023 & \bf0.128 & \bf7.776 & 2.441 & \bf0.720 & 0.720 \\
\bottomrule
\label{tab:baseline-comparisons}
\end{tabular}

\end{table*}

\begin{table}[!htp]
\centering
\caption{Comparison of SSP-GNN vs Edge Belief on scenarios of diverse combinations of false alarms (FA) and ReID Strength}
\sisetup{detect-all}
\NewDocumentCommand{\B}{}{\fontseries{b}\selectfont}
\begin{tabular}{
  l
  S[table-format=1.2]
  S[table-format=1.2]
  S[table-format=1.2]
}
\toprule
Tracker & {FA Rate} & {ReID Strength} & {MOTA $\uparrow$}\\
\midrule
\rowcolor{gray!20}
Edge-Belief & 10 &\text{Moderate} & \text{0.862 $\pm$ 0.012}\\
\rowcolor{gray!20}SSP-GNN & 10 & \text{Moderate} & \textbf{0.950 $\pm$ 0.019}\\  
Edge-Belief & 10 & \text{Weak} & \text{0.826 $\pm$ 0.008}\\
SSP-GNN & 10 & \text{Weak} & \textbf{0.952 $\pm$ 0.006}\\
\rowcolor{gray!20} Edge-Belief & 20 & \text{Moderate} & \text{0.656 $\pm$ 0.032}\\ 
\rowcolor{gray!20}SSP-GNN & 20 &\text{Moderate} & \textbf{0.891 $\pm$ 0.034}\\
Edge-Belief & 20 & \text{Weak} & \text{0.633 $\pm$ 0.018}\\
SSP-GNN & 20 & \text{Weak} & \textbf{0.873 $\pm$ 0.023}\\
\bottomrule
\label{tab:edgebelief_vs_ssp}
\end{tabular}
\end{table}

\vxapara{Effect of ReID feature strength and noisy  dimensions} We use one trajectory following an Ornstein–Uhlenbeck motion model over a 50 timestep period with a high false alarm rate (expected 30 per frame). Table~\ref{tab:reID-strength} shows a clear positive correlation between model performance and ReID feature strength. The two extremes of the table also indicate that the model leverages strong ReID features effectively, while also allowing for good motion-based predictions when ReID features are highly unreliable.

While it is clear that our method can handle higher-dimensional ReID features, it is interesting to consider a case where the dimensions are not equally informative. We therefore constructed scenarios with up to ten additional ReID feature dimensions filled with i.i.d. noise. While training took longer, our learned tracker did not suffer any noticeable degradation in accuracy, indicating that the algorithm is able to find informative components of ReID and ignore the rest.

\begin{table}[!htp]
\centering
\caption{Effect of decreasing the strength of ReID features}
\sisetup{detect-all}
\NewDocumentCommand{\B}{}{\fontseries{b}\selectfont}
\begin{tabular}{
  @{}
  l
  S[table-format=1.2]
  S[table-format=1.2]
  S[table-format=1.2]
  S[table-format=1.2]
  S[table-format=1.2]
  @{}
}
\toprule
ReID Str. & {MOTA $\uparrow$} & {FP Rate $\downarrow$} & {FN Rate $\downarrow$} & {IDS Rate $\downarrow$} \\
\midrule
Strong & \text{0.961 $\pm$ 0.031} & 0.015 & 0.025 & 0.000\\ 
Moderate & \text{0.918 $\pm$ 0.023} & 0.041 & 0.041 & 0.000\\
Weak & \text{0.909 $\pm$ 0.009}  & 0.045 & 0.045 & 0.000\\
Very weak & \text{0.904 $\pm$ 0.021} & 0.048 & 0.048 & 0.000\\
\bottomrule
\label{tab:reID-strength}
\end{tabular}

\end{table}

Table~\ref{tab:reID-noisy-dimensions} demonstrates that the model is resilient to added dimensions of noise in the ReID features. There is a very slight drop in prediction quality as initial noisy dimensions are added, but then a sudden increase at 10. We attribute this to a random effect, although there could be an unforeseen interaction where adding extra noise incentivizes the model to learn the meaningful relations more robustly on our limited training set.

\vxacut{
\begin{table*}[!htp]
\label{reID-strength-features}
\centering
\caption{Evaluation of Predicted Tracks for varying strength of ReID features. NOTE: current table contents do not demonstrate ReID usefulness, harder scenario should probably be used.}
\sisetup{detect-all}
\NewDocumentCommand{\B}{}{\fontseries{b}\selectfont}
\begin{tabular}{
  @{}
  l
  S[table-format=1.2]
  S[table-format=1.2]
  S[table-format=1.2]
  S[table-format=1.2]
  S[table-format=1.2]
  S[table-format=1.2]
  S[table-format=1.2]
  S[table-format=1.2]
  S[table-format=1.2]
  S[table-format=1.2]
  S[table-format=1.2]
  S[table-format=1.2]
  @{}
}
\toprule
ReID Strength & \multicolumn{11}{c}{Metrics} \\
\cmidrule(l){2-12}
& \multicolumn{3}{c}{MOTA} & \multicolumn{4}{c}{SIAP} & \multicolumn{4}{c}{GOSPA} \\
\cmidrule(lr){2-4} \cmidrule(lr){5-8} \cmidrule(lr){9-12}
& {FP Rate $\downarrow$} & {FN Rate $\downarrow$} & {MOTA $\uparrow$} &{Comp. (1)} & {Amb. (1)}& {Spur. (0)} & {Pos. Error. (0)} & {Dist. $\downarrow$} & {Local. $\downarrow$} & {Missed $\downarrow$} & {False $\downarrow$}\\
\midrule
Very Weak & 0.076 & 0.076 & 0.849 & 0.693 & 1.000 & 0.308 & 0.005 & 6.634 & 11.817 & 0.720 & 0.720 \\
Weak & 0.026 & 0.026 &  0.949 & 0.733  & 1.000 & 0.267 & 0.001 & 2.126 & 3.964 & 0 & 0 \\
Moderate & 0.013 & 0.013 & 0.974 & 0.707 & 1.000 & 0.293 & 0 & 1.196 & 2.277 & 0 & 0 \\
Strong & 0. & 0 & 1 & 1 & 1 & 0 & 0 & 0 & 0 & 0 & 0\\ 
\bottomrule
\end{tabular}

\end{table*}}

\begin{table}[!htp]
\centering
\caption{Effect of additional noisy ReID dimensions}
\sisetup{detect-all}
\NewDocumentCommand{\B}{}{\fontseries{b}\selectfont}
\begin{tabular}{
  @{}
  l
  S[table-format=1.2]
  S[table-format=1.2]
  S[table-format=1.2]
  S[table-format=1.2]
  S[table-format=1.2]
  @{}
}
\toprule
Noisy Dims & {MOTA $\uparrow$} & {FP Rate $\downarrow$} & {FN Rate $\downarrow$} & {IDS Rate $\downarrow$}\\
\midrule
0 & \text{0.935 $\pm$ 0.021} & 0.033 & 0.033 & 0.000\\
2 & \text{0.929 $\pm$ 0.010} & 0.036 & 0.036 & 0.000\\
5 & \text{0.927 $\pm$ 0.011} & 0.037 & 0.037 & 0.000\\
10 & \text{0.959 $\pm$ 0.015} & 0.020 & 0.020 & 0.000\\
\bottomrule
\label{tab:reID-noisy-dimensions}
\end{tabular}

\end{table}

\begin{table}[!htp]
\centering
\caption{Effect of increasing the rate of false alarms}
\sisetup{detect-all}
\NewDocumentCommand{\B}{}{\fontseries{b}\selectfont}
\begin{tabular}{
  @{}
  l
  S[table-format=1.2]
  S[table-format=1.2]
  S[table-format=1.2]
  S[table-format=1.2]
  S[table-format=1.2]
  @{}
}
\toprule
FA per frame & {MOTA $\uparrow$} & {FP Rate $\downarrow$} & {FN Rate $\downarrow$} & {IDS Rate $\downarrow$} \\
\midrule
2 & \text{0.994 $\pm$ 0.004} & 0.003 & 0.003 & 0.001 \\
5 & \text{0.979 $\pm$ 0.007}  & 0.009 & 0.009 & 0.004 \\
10 & \text{0.938 $\pm$ 0.047} & 0.029 & 0.029 & 0.005 \\
20  & \text{0.900 $\pm$ 0.040} & 0.048 & 0.046 & 0.006 \\ 
30 & \text{0.907 $\pm$ 0.017} & 0.045 & 0.046 & 0.003\\ 
\bottomrule
\label{tab:fa_freq}
\end{tabular}

\end{table}

\vxapara{Effect of false alarm rate} Table~\ref{tab:fa_freq} displays the performance of our method at a variety of expected numbers of false alarms per frame. The performance is nearly optimal when false alarms are sparse, and deteriorates gracefully as false alarms are added, however it seems to level off and not drop below a MOTA score of about 0.9. This indicates that our algorithm is capable of capitalizing on easier regimes with relatively few false positives, while also adapting to hard environments where false alarms significantly outnumber ground truth tracks.

\vxacut{
Potential Extra Table for SSP GNN vs Edge Beleif comparison. each plot has fixed ReID strength (strong, medium, weak) and we test with 4 different False Alarm Rate?
\begin{table}[!htp]
\centering
\caption{SSP GNN vs Edge Belief against scenarios of various False Alarm Rates (in parentheses)}
\sisetup{detect-all}
\NewDocumentCommand{\B}{}{\fontseries{b}\selectfont}
\begin{tabular}{
  @{}
  l
  S[table-format=1.2]
  S[table-format=1.2]
  S[table-format=1.2]
  S[table-format=1.2]
  S[table-format=1.2]
  @{}
}
\toprule
Tracker & {MOTA $\uparrow$} & {FP Rate $\downarrow$} & {FN Rate $\downarrow$} & {IDS Rate $\downarrow$} \\
\midrule
Edge Belief (2) & \text{0.000 $\pm$ 0.000} & 0.000 & 0.000 & 0.000\\
SSP-GNN (2) & \text{0.000 $\pm$ 0.000} & 0.000 & 0.000 & 0.000\\
\midrule
Edge Belief (5) & \text{0.000 $\pm$ 0.000} & 0.000 & 0.000 & 0.000\\
SSP-GNN (5) & \text{0.000 $\pm$ 0.000} & 0.000 & 0.000 & 0.000\\
\midrule
Edge Belief (10) & \text{0.000 $\pm$ 0.000} & 0.000 & 0.000 & 0.000\\
SSP-GNN (10) & \text{0.000 $\pm$ 0.000} & 0.000 & 0.000 & 0.000\\
\midrule
Edge Belief (20) & \text{0.000 $\pm$ 0.000} & 0.000 & 0.000 & 0.000\\
SSP-GNN (20) & \text{0.000 $\pm$ 0.000} & 0.000 & 0.000 & 0.000\\
\bottomrule
\label{tab:reID-noisy-dimensions}
\end{tabular}

\end{table}
}

\vxapara{Effect of the training set size} Table~\ref{tab:training-set-size} tests our model with a variety of training set sizes. Our experiments use limited size training sets so as to run multiple trials. Real world applications would likely train on more data than we use, so it is important that model accuracy  scales with training set size.

This experiment uses a diverse array of scenarios using constant velocity and Ornstein–Uhlenbeck motion models with variable ReID strength (moderate to weak) and up to 5 ground truth tracks. Table~\ref{tab:training-set-size} shows a positive correlation between number of graphs in the training set and model performance. Learned tracker consistency, represented by standard deviation, also improves substantially with training set size.

\begin{table}[!htp]
\centering
\caption{Effect of increasing training set size}
\sisetup{detect-all}
\NewDocumentCommand{\B}{}{\fontseries{b}\selectfont}
\begin{tabular}{
  @{}
  l
  S[table-format=1.2]
  S[table-format=1.2]
  S[table-format=1.2]
  S[table-format=1.2]
  S[table-format=1.2]
  @{}
}
\toprule
Train Set\\Size & {MOTA $\uparrow$} & {FP Rate $\downarrow$} & {FN Rate $\downarrow$} & {IDS Rate $\downarrow$} \\
\midrule
1 & \text{0.894 $\pm$ 0.046} & 0.045 & 0.053 & 0.007\\
5 & \text{0.937 $\pm$ 0.032} & 0.024 & 0.032 & 0.007\\
10 & \text{0.954 $\pm$ 0.006} & 0.020 & 0.020 & 0.007\\
20 & \text{0.958 $\pm$ 0.019} & 0.016 & 0.016 & 0.011\\ 
30 & \text{0.963 $\pm$ 0.007} & 0.014 & 0.014 & 0.009\\ 
\bottomrule
\label{tab:training-set-size}
\end{tabular}

\end{table}

\vxapara{Effect of lower detection probability} Finally, we consider the effect of an imperfect detector i.e., $P_D < 1.0$. To allow the model to select trajectories that skip time steps, we modify the tracking graph to contain edges between non-adjacent timesteps up to three apart.

In Table~\ref{tab:p_d_exps} we observe a drop in MOTA of around 0.02 when detection probability decreases from 1.0 to 0.95, as adding missed detections increases the complexity of the representation that must be learned. Our model performance is resilient between detection probabilities of 0.95 to 0.85. From 0.85 to 0.8, there is another drop in performance, which may be due to the increasing prevalence of enduring missed detections lasting beyond what the model is capable of considering.

\begin{table}[!htp]
\centering
\caption{Effect of decreasing $P_D$}
\NewDocumentCommand{\B}{}{\fontseries{b}\selectfont}
\begin{tabular}{
  @{}
  l
  S[table-format=1.2]
  S[table-format=1.2]
  S[table-format=1.2]
  S[table-format=1.2]
  S[table-format=1.2]
  S[table-format=1.2]
  @{}
}
\toprule
$P_D$ & {MOTA $\uparrow$} & {FP Rate $\downarrow$} & {FN Rate $\downarrow$} & {IDS Rate $\downarrow$} \\
\midrule
1.0 & \text{0.978 $\pm$ 0.001} & 0.010 & 0.010 & 0.003 \\
0.95 & \text{0.957 $\pm$ 0.013} & 0.022 & 0.014 & 0.007 \\
0.9 & \text{0.945 $\pm$ 0.020} & 0.024 & 0.026 & 0.004 \\
0.85 & \text{0.944 $\pm$ 0.012} & 0.031 & 0.022 & 0.002 \\
0.8 & \text{0.913 $\pm$ 0.017} & 0.048 & 0.034 & 0.006 \\
\bottomrule
\label{tab:p_d_exps}
\end{tabular}
\end{table}

\section{Conclusions}
We presented a novel trainable approach for graph-based multi-object tracking. Our approach predicts edge costs using a message passing GNN, then relies on a globally optimal and computationally-efficient path-finding algorithm to form tracks that satisfy the required constraints. The message-passing network can be learned end-to-end in a bilevel optimization framework that is different from prior work. We thoroughly investigated model performance on a variety of scenarios. Our learned tracker compares favorably to a strong baseline. In our future work, we plan to focus on computationally-efficient extensions to our graph formulation and the message-passing network to optimally handle long-term occlusions.

\vxacut{Our experiments demonstrate that training exhibits stable convergence that matches the complexity of the training scenarios.}

\vspace{-0.05in}
\section{Acknowledgements}
We thank the anonymous reviewers for their feedback and Samuel C. Buckner for helpful discussions. Earlier work funded by U.S. Office of Naval Research Code 321.

\bibliographystyle{ieee}
\bibliography{uw_gnn_mot_fusion2024, uw_gnn_mot}

\end{document}